\documentclass{article}
\usepackage{spconf,amsmath,graphicx}

\usepackage{booktabs}
\usepackage{algorithm}
\usepackage{algorithmic}

\usepackage{listings}
\usepackage[dvipsnames]{xcolor}
\usepackage{color}
\definecolor{deepblue}{rgb}{0,0,0.5}
\definecolor{deepred}{rgb}{0.6,0,0}
\definecolor{deepgreen}{rgb}{0,0.5,0}
\definecolor{codebrown}{rgb}{0.8,0.44,0.2}
\definecolor{codegray}{rgb}{0.5,0.5,0.5}
\definecolor{codepurple}{rgb}{0.58,0,0.82}
\definecolor{backcolour}{rgb}{0.95,0.95,0.92}

\usepackage{multirow}

\def\ourmethod{TVT}

\title{\ourmethod{}: Training-free Vision Transformer Search on Tiny Datasets}
\name{Zimian Wei$^{1}$,  Hengyue Pan$^{1}$, Lujun Li$^{2}$, Peijie Dong$^{2}$, Zhiliang Tian$^{1}$, Xin Niu$^{1}$,  Dongsheng Li$^{1}$}
\address{$^1$ College of Computer, National University of Defense Technology\\
 $^2$ The Hong Kong University of Science and Technology\\
 }

\begin{document}
%
\maketitle
\begin{abstract}
Training-free Vision Transformer (ViT) architecture search is presented to search for a better ViT with zero-cost proxies. While ViTs achieve significant distillation gains from CNN teacher models on small datasets,  the current zero-cost proxies in ViTs do not generalize well to the distillation training paradigm according to our experimental observations.
In this paper, for the first time, we investigate how to search in a training-free manner with the help of teacher models and devise an effective Training-free  ViT (TVT) search framework. 
Firstly, we observe that the similarity of attention maps between ViT with ConvNet teachers affects distill accuracy notably. Thus, we present a teacher-aware metric conditioned on the feature attention relations between teacher and student. 
Additionally, TVT employs $L_2$-norm of the student's weights as the student-capability metric to improve ranking consistency. 
Finally, TVT searches for the best ViT for distilling with ConvNet teachers via our teacher-aware metric and student-capability metric, resulting in impressive gains in efficiency and effectiveness. Extensive experiments on various tiny datasets and search spaces show that our TVT outperforms state-of-the-art training-free search methods. The code will be released.
\end{abstract}
\section{Introduction}
\label{sec:intro}

ViTs (ViTs) have achieved remarkable performance in the computer vision field.
However, there are still two fundamental problems that seriously limit the broad application of the family of ViT models: \textbf{large-scale training data requirements and enormous model sizes.} 
ViTs lack some priori properties in ConvNets, such as locality, weight-sharing mechanisms, and translation equivalence ~\cite{localvit}. 
As a result, without the pre-training stage on a large dataset,  the performance of ViTs is relatively weak. 
Additionally, ViT models generally need more parameters and are harder to train. 

\begin{figure}[t]
    \centering
    \includegraphics[width=0.9\linewidth]{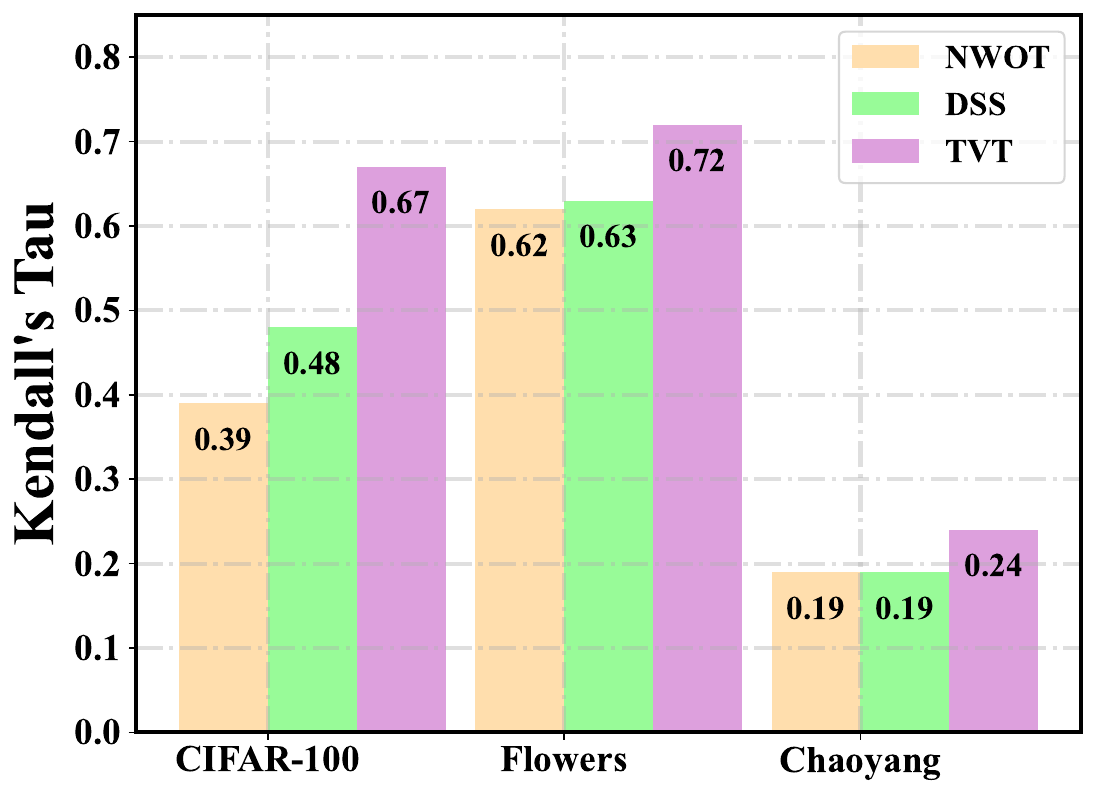}
    \vspace{-1em}
    \caption{The rank consistency of Zero-cost proxies on the three tiny datasets of CIFAR100, Flowers and Chaoyang. Results indicated that our proposed \ourmethod{} significantly outperformed NWOT~\cite{NWOT} and DSS~\cite{DSS}.}
    \label{fig:accuracy}
\end{figure}

To address the issue of data inefficiency, many attempts~\cite{deit,localvit,Jia2021EfficientVT,Wu2022TinyViTFP} leverages Knowledge Distillation (KD)~\cite{liu2023norm,li2023auto,lishadow,li2023kd,li2022tf,li2022SFF,li2020explicit,li2022self,shao2023catch} to enhance the data efficiency of ViTs by using ConvNets as teachers. 
With the inductive biases contained in the dark knowledge from teachers, ViTs obtain significant improvements in both accuracy performance and convergence speed.
For example, DeiT~\cite{deit} uses a distill token to develop the data efficiency of training on ImageNet with distilling knowledge from ConvNets. 
LG~\cite{li2022locality} involves locality guidance via distilling from lightweight ConvNets and improves the accuracy of various ViTs on tiny datasets with $13.07 \%\sim 7.85\%$ margins. 
Therefore, employing ConvNets for knowledge distillation has become a widely recognized and effective approach to achieving data-efficient ViTs on small datasets.

\begin{figure*}[t]
    \centering
    \includegraphics[width=1.0\linewidth]{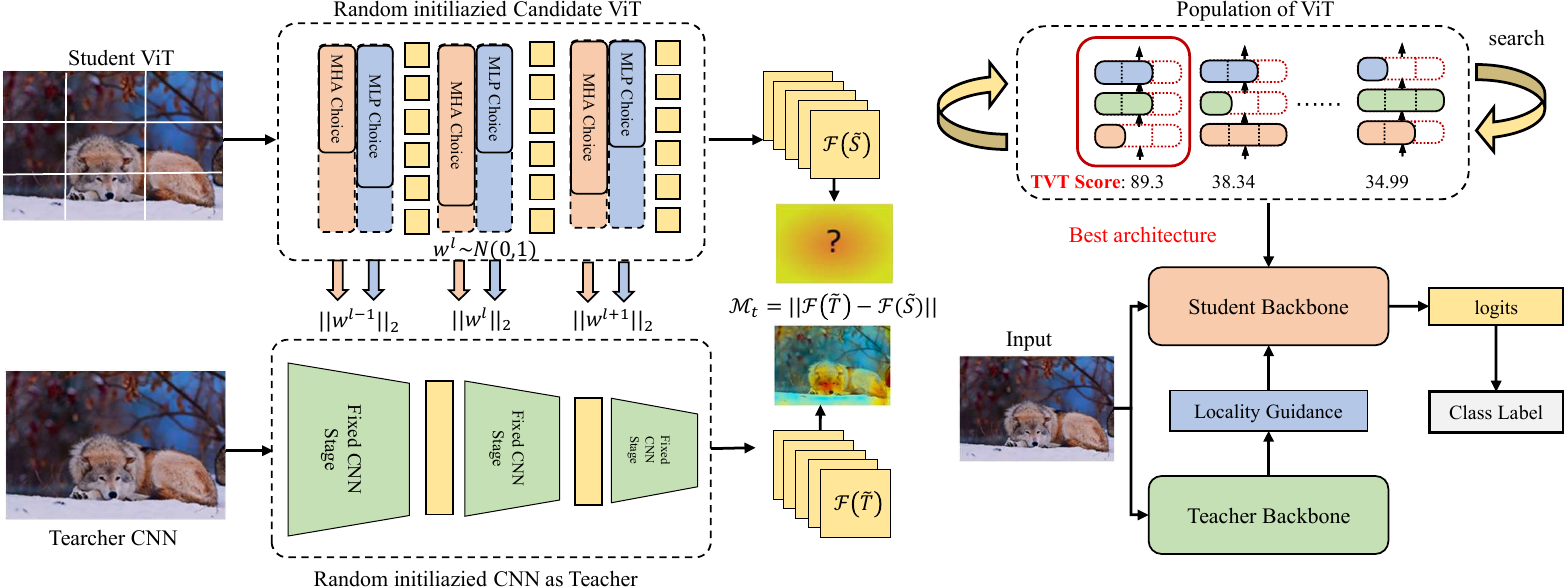}
    \caption{Illustration of the proposed method. \textbf{Left}: The calculation of the proposed \ourmethod{} Zero-cost Proxy consists of Teacher-Aware Metric $\mathcal{M}_{t}$ and student-capability Metric $\mathcal{M}_{s}$.
        $\mathcal{M}_{t}$ measures the gap between the student and ConvNet Teacher, with the motivation that more potential ViTs can be selected by finding those with an attention map closer to the teacher network.
        $\mathcal{M}_{s}$ is utilized to assess the potential of the candidate network by quantifying the magnitude of its weights. \textbf{Right}: The searching and distillation pipeline.
        Upon obtaining the best architecture with the highest TVT score, a distillation process with locality guidance~\cite{li2022locality} and logit activation is conducted to train the searched student ViT.}
    \label{fig:main}
    \label{-6mm}
\end{figure*}

For the problem of model redundancy, Neural Architecture Search (NAS) approaches~\cite{wei2023Auto-Prox,li2021nas,linas2,lichengp,dong2023diswot,dong2023emq} improve model efficiency by building a search space and searching for optimal candidate ViTs. 
In contrast to traditional training-based NAS with huge overhead in supernet training, recent training-free NAS has received great research interest owing to their extremely low cost. 
These methods exploit zero-cost proxies~\cite{syflow,TENAS} to predict the actual accuracy ranking of candidate architectures. 
Conditional on model internal information, zero-cost proxies use just a single minibatch of training data to compute a proxy score without training. 
However, existing proxies do not generalize well in distilling scenarios (see Figure~\ref{fig:accuracy}) due to their lack of external information from teachers.
To this end, there is a need to consider both external information from teachers and the architectural designs for zero-cost proxies.


Inspired by the above analysis and observations, we present TVT, a simple yet effective training-free ViTs architecture search framework under distillation scenarios on tiny datasets. 
Specifically, we design a zero-cost proxy consisting of a teacher-aware metric and a student-capability metric.
The teacher-aware metric is based on the motivation that the student network with a smaller teacher-student gap can result in superior distillation performance.
We utilize spatial attention maps from ViT (student) and ConvNet (teacher) for higher-level representation information, and then take their $L_2$ distance as the teacher-aware metric (see Figure~\ref{fig:main}).
As for the student-aware metric, the capacity of the student model also shows a non-trivial impact on the distillation accuracy, which means the ability to represent more complex functions.
We use the L2-norm of model parameters, which is largely related to the model's capacity.
Based on the designed zero-cost proxy, we search for optimal student architectures and then implement the distillation process between the searched ViT student and the pre-defined ConvNet teacher.


Our TVT method delivers two-fold merits: 
(1)~\textbf{Efficiency.}  In contrast to training-based architecture search methods, such as EcoNAS~\cite{EcoNAS} and ENAS~\cite{ENAS}, which require individual or weight-sharing training, TVT does not necessitate the training of student models. 
Instead, TVT evaluates models by the designed zero-cost proxy, which only requires forward calculation at initialization. 
(2) \textbf{Effectiveness.}  Our proposed TVT proxy is effective in improving accuracy due to its adaptation for distillation. 
With the teacher-aware metric and student-capability metric, our TVT surpasses existing training-free NAS approaches by a large margin (see Figures~\ref{fig:accuracy}). 


We conduct extensive experiments on CIFAR-100, Flowers, and Chaoyang~\cite{zhu2021hard} datasets.
The experiments demonstrate that TVT can achieve better distillation accuracy and superior rank consistency than other methods.
We also conduct comprehensive ablation studies to investigate the effectiveness of different designs in our method.

Our main contributions are as follows: 
1) Our TVT is the first work focusing on a training-free search for ViTs on tiny datasets under distillation.  
2) Our TVT searches for the optimal ViT for distilling with ConvNet teachers, leading to substantial gains in efficiency and effectiveness.
3) We experimentally validate that TVT achieves state-of-the-art performances in multiple datasets and search spaces and can advance the wider application of transformers in vision tasks.

\section{Methodology}
\label{section:method}


An overview of the proposed \ourmethod{} is presented in Figure \ref{fig:main}. 
The main components of \ourmethod{} include the design of zero-cost proxy and optimal student network search. 
We give the details of these two designs in the following sections.

\subsection{\ourmethod{} Zero-cost Proxy}
The \ourmethod{} zero-cost proxy consists of two components, including the teacher-aware metric and student-capability metric. 

\noindent \textbf{Teacher-aware Metric.} 
Teacher-aware metric focuses on reflecting the gap between the teacher and student networks, which helps recognize the optimal student ViT with a minimal teacher-student gap. 
To achieve this, we utilize spatial attention maps, which provide higher-level semantic information by revealing spatial areas of the inputs that the network focuses most for decision.

For the input image $ X\in \mathbf{R}^{3\times H\times W}$, the feature map of the teacher network is $T\in \mathbf{R}^{C_T\times H\times W}$, where $C_T$ is the output number of channels and $H \times W$ is the spatial size. For a random vision transformer $S_i$, the token sequence after the encoder block is ${S}^{i}\in \mathbf{R}^{ C_i\times L}$, which can be flattened as ${S}^{i}\in \mathbf{R}^{ C_i\times P\times P}$. $L$ is the number of tokens, $P$ is the patch size, and $ C_i $ is the embedding dimension.

We introduce a mapping function $\mathcal{F}$, which generates a spatial attention map by computing statistics across the channel dimension.
Before input to the mapping function $\mathcal{F}$, $T$ is interpolated to the same spatial size as ${S}^{i}$, resulting a new feature map $\tilde{T}\in \mathbf{R}^{ C_T\times P\times P}$.
Then the generated attention matrix of the teacher and student network is formulated as follows:
\begin{align}
    \mathcal{F}(\tilde{T})=\phi(\sum^{C_T}_{n=1} \tilde{T}_n^2), \quad
    \mathcal{F}({S}^{i})=\phi(\sum^{C_i}_{n=1}{{S}^{i}_n}^2)
\end{align}
\noindent where $\phi$ is a normalization function, $n$ is the index of channel dimension. Thus $\tilde{T_n}=\tilde{T}(n,:,:)$ and ${S}^{i}_n={S}^{i}(n,:,:)$.
Then, the $\mathcal{M}_{t}$ for a candidate ViT $S^i$ is as follows:
\begin{equation}
    \label{sample_metric}
    \mathcal{M}_{t}=\left\|\mathcal{F}(\tilde{T})- \mathcal{F}({S}^{i})\right\| _2.
\end{equation}

\noindent \textbf{Student-capability Metric.} 
We introduce the student-capability metric to reveal the potential of ViTs, which is also closely related to the distillation performance.
As studied in pruning-based zero-cost proxies \cite{zeronas,snip}, the number of important weight parameters correlates positively with the model capacity. 
It motivates us to sum over the $L_2$-norms of parameters in each layer to score a ViT. 
Given a candidate ViT architecture, the student-capability metric can be formulated as follows:
\begin{equation}
    \begin{aligned}
        \mathcal{M}_{s}=\sum _{n} \left\| W_n \right\| _2.
    \end{aligned}\label{q3}\end{equation}

\noindent where n is the index of building blocks in a candidate ViT. 

The student-capability metric is related to weight initialization and parameters.
For a fair comparison, we use the same weight initialization for all zero-cost proxies.
Regarding parameters, a student ViT with more parameters generally has a higher capacity but does not always lead to higher distillation accuracy due to over-fitting and the teacher-student gap.
To this end, we formulated the \ourmethod{} proxy score as:
\begin{equation}
    \mathcal{M}_{\ourmethod{}} =  \alpha \times f(\mathcal{M}_{s}) + \beta \times f(\mathcal{M}_{t}).
    \label{eq:tvt}
\end{equation}

\noindent where $\alpha$, $\beta$ are hyper-parameters, and $f$ is a min-max normalization method.

\subsection{Training-free Vision Transformer Search}
 We conduct a training-free ViT search to efficiently discover the optimal student $\alpha^* $ from  search space $\mathcal{A}$, which is formulated as:
\begin{equation}
    \alpha^* =  \arg\max_{\alpha \in \mathcal{A}}(\mathcal{M}_{\ourmethod{}}).
\end{equation}where the argmax function is applied to find an architecture that maximizes the TVT proxy score. 
Since gradient back-propagation is not utilized, our search algorithm is demonstrated to be effective. We present the procedure for discovering the optimal student in Algorithm \ref{alg:evolution}.

\begin{algorithm}[t]
    \small
    \caption{Training-free ViT Search with \ourmethod{} proxy score}
    \label{alg:evolution}
    \textbf{Input}: Search space $\mathcal{S}$, population size $\mathcal{N}$,  topk $k$, teacher network $\mathcal{T}$.

    \textbf{Output}: \leftline{ViT with Top-1 TVT proxy score.}
    \begin{algorithmic}[1]
        \STATE $\mathcal{P}$ := Random sample population$(\mathcal{N},\mathcal{S})$;
        \FOR{$ Candidates \{A_i\} \in \mathcal{P}$}
        \STATE  Get $\mathcal{M}_{t}$($Candidates \{A_i\}, \mathcal{T}$);
        \STATE  Get $\mathcal{M}_{s}$($Candidates \{A_i\}$);
        \STATE Get TVT-Score $z$ = $\mathcal{M}_{\ourmethod{}}(\mathcal{M}_{s}, \mathcal{M}_{t}$);
        \STATE update topk TVT proxy score $k$ ;

        \ENDFOR
    \end{algorithmic}
\end{algorithm}


\section{Experiments}
\label{sec:typestyle}

\subsection{Implementation Details:} 
Our method consists of a training-free ViT search process and a distillation training process for the searched ViT.
During the search process, we compute the TVT proxy score for each candidate ViT.
ViTs are sampled from the AutoFormer-Ti~\cite{chen2021autoformer} and PiT~\cite{DSS} search spaces, with parameter intervals of $4\sim9$ M and $2\sim25$ M.
We set $\alpha$ and $\beta$ in Equation (4) as $2$ and $-3$, respectively.
Without gradient back-propagation, the time cost of \ourmethod{} to search among 1,000 candidate ViTs is around $1$ hour on a single A40 GPU. 
After the search process is completed, we train the obtained ViT with both classification loss and distillation loss. 
We follow the training hyper-parameters in \cite{li2022locality}.
When evaluating various zero proxies, we randomly select 100 ViTs and compute the  Kendall ranking correlation between their actual distillation accuracy and TVT proxy scores.
We repeat three runs and report the average value for each rank correlation result.





\vspace{-1em}


\begin{table}[t]
    \label{tab:nas}
    \centering
        \caption{Vanilla and distillation results (\%) when searching ViT from AutoFormer search space with different proxies. Vanilla means vanilla classification accuracy. Distill Acc represents the final accuracy of ViTs under distilling training~\cite{li2022locality}.
        }
    \footnotesize
    \begin{tabular}{c|c|c|c|c}
        \toprule
        Dataset & Proxy & Teacher     & Vanilla Acc.                                                                                           & Distill Acc.   \\

        \midrule
        \multirow{6}{*} {CIFAR-100}
                                      & Vanilla       & \multicolumn{1}{c|}{\multirow{6}{*}{71.16}}           & 69.69                                                    &  75.08          \\
                                      & GraSP \cite{grasp}   & \multicolumn{1}{c|}{}    & 69.54                                                                             & 75.51          \\
                                      & TE-NAS~\cite{TENAS}   & \multicolumn{1}{c|}{}   & 69.51                                                                              & 75.70          \\
                                      & NWOT~\cite{NWOT}   & \multicolumn{1}{c|}{}          & 68.00                                                                      & 78.12          \\
                                      & DSS~\cite{DSS}    & \multicolumn{1}{c|}{}             & 67.86                                                                     & 78.32          \\
                                      & \ourmethod{}(Ours)  & \multicolumn{1}{c|}{}     & 68.82                                                                & \textbf{78.80} \\
        \midrule
        \multirow{6}{*} {Flowers}
                                      & Vanilla      & \multicolumn{1}{c|}{\multirow{6}{*}{59.38}}         & 54.98                                                       &    67.96          \\
                                      & GraSP \cite{grasp}   & \multicolumn{1}{c|}{}            & 53.65                                                                    & 66.37          \\
                                      & TE-NAS~\cite{TENAS}   & \multicolumn{1}{c|}{}           & 54.07                                                                     & 67.26          \\
                                      & NWOT~\cite{NWOT}   & \multicolumn{1}{c|}{}             & 53.41                                                                     & 68.69          \\
                                      & DSS~\cite{DSS}     & \multicolumn{1}{c|}{}            & 54.29                                                                     & 68.63          \\
                                      & \ourmethod{}(Ours)  & \multicolumn{1}{c|}{}   & 53.31                                                                   & \textbf{69.13} \\
        \midrule
        \multirow{6}{*} {Chaoyang}
                                      & Vanilla     &   \multicolumn{1}{c|}{\multirow{6}{*}{79.66}}             &  82.84                          &   85.04       \\
                                      & GraSP \cite{grasp}  & 
                                      \multicolumn{1}{c|}{}       &  81.67                                                                    &  84.90         \\
                                      & TE-NAS~\cite{TENAS}   & \multicolumn{1}{c|}{}    &  81.39                                                                   &   84.67        \\
                                      & NWOT~\cite{NWOT}    & \multicolumn{1}{c|}{}     & 81.77                                                                 &    85.13       \\
                                      & DSS~\cite{DSS}    & \multicolumn{1}{c|}{}      &  82.75                                                                   &  85.13         \\
                                      & \ourmethod{}(Ours)  & \multicolumn{1}{c|}{}     &  81.44                                                                   & \textbf{85.83} \\
        \bottomrule
    \end{tabular}
    \label{tab:zc_search_acc}
\end{table}

\begin{table}[t]
    \centering
        \caption{ Kendall ranking correlation between distillation accuracy and \ourmethod{} proxy score of 100 randomly sampled ViTs.     
        }
    \footnotesize
    \begin{tabular}{c|c|cc}
        \toprule
        \multirow{2}{*}{Datasets} & \multirow{2}{*}{Proxy} & \multicolumn{2}{c}{Search Space}                                            \\
                                  &                                                  &  AutoFormer                    &  PiT \\  \midrule
        \multirow{5}{*}{CIFAR-100}
                                  & GraSP \cite{grasp}                                 & -0.39                       & -0.28     \\
                                  & TE-NAS \cite{TENAS}                                 & -0.39                          & -0.16     \\
                                  & NWOT  \cite{NWOT}                                  & 0.60                           & 0.46      \\
                                  & DSS \cite{DSS}                                     & 0.48                         & 0.74        \\
                                  & \ourmethod{} (Ours)                                  & \textbf{0.67}                          & \textbf{0.76}       \\ \midrule
        \multirow{5}{*}{Flowers}
                                  & GraSP \cite{grasp}                                 & -0.62                       & -0.12     \\
                                  & TE-NAS \cite{TENAS}                                 & -0.58                         & -0.15      \\
                                  & NWOT  \cite{NWOT}                                    & 0.68                          & 0.57      \\
                                  & DSS \cite{DSS}                                      & 0.63                           & 0.72       \\
                                  & \ourmethod{}(Ours)                                   & \textbf{0.72}                         & \textbf{0.84}      \\ \midrule
        \multirow{5}{*}{Chaoyang}
                                  & GraSP \cite{grasp}                                 & -0.19                     & -0.11     \\
                                  & TE-NAS \cite{TENAS}                                 & -0.26                        & -0.14    \\
                                  & NWOT  \cite{NWOT}                                  & 0.18                          & 0.52        \\
                                  & DSS \cite{DSS}                                      & 0.19                         & 0.42      \\
                                  & \ourmethod{} (Ours)                       & \textbf{0.24}                        & \textbf{0.58}       \\ \bottomrule
    \end{tabular}
    \label{tab:rank_consist}
\end{table}

\subsection{Experimental Results}
Table~\ref{tab:zc_search_acc} presents the results of the vanilla and distillation models when searching with different zero-cost proxies on the AutoFormer search space. The insights from the results of Table~\ref{tab:zc_search_acc} can be summarized in three points. 
Firstly, among the presented zero-cost proxies, \ourmethod{} consistently yields better distillation performance, indicating that \ourmethod{} is more precise in predicting distillation accuracy. Secondly, ViTs with higher vanilla accuracy does not necessarily result in better distillation results. We speculate that this may be due to the semantic information gap between the ConvNet teacher and student ViTs. Nevertheless, a poorer teacher can still provide promising guidance for the training of student ViTs on tiny datasets. 
This phenomenon is also demonstrated in LG~\cite{li2022locality}.
Lastly, ViTs can obtain substantial gains from knowledge distillation (up to 10\% accuracy), surpassing both vanilla and teacher results. It demonstrates the large potential of ViTs to surpass ConvNet on tiny datasets. 
In addition to the distillation accuracy, we also present the rank consistency of zero-cost proxies under different experimental settings in Table~\ref{tab:rank_consist}. 
Results show that \ourmethod{} significantly outperforms other excellent zero-cost proxies, demonstrating its effectiveness.

\begin{table}[ht]
    \centering
        \caption{Ablation on the design  of teacher-aware metric $\mathcal{M}_{t}$.}
	 \resizebox{1\linewidth}{!}{
    \begin{tabular}{lccc}
        \toprule
        Method  & Sample Relation   & MMD  Metric  &  Attention Map  \\
        \midrule
        Kendall & 0.40 & 0.38 & \textbf{0.46} \\
        \bottomrule
    \end{tabular}%
    }

    \label{tab:transform}

\end{table}%

\begin{table}[!ht]
    \centering
    \small
        \caption{Ablation on hyper-parameters in Equation~\ref{eq:tvt}.}\label{tab:beta}

    \begin{tabular}{ccc|ccc}
        \toprule
        $ \alpha $ & $\beta$ & Kendall & $ \alpha $ & $\beta$ & Kendall       \\
        \midrule
        1         & -1       & 0.44    & 2         & -1       & -0.11         \\
        1         & -2       & 0.64    & 2         & -2       & 0.44          \\
            0         & -1       & 0.46    & 1         & 0       & 0.60          \\
        1         & -3       & 0.56    & 2         & -3       & \textbf{0.67} \\
        \bottomrule
    \end{tabular}

\end{table}%

      
        


\subsection{Ablation Study}
\label{section:ablation}

To shed light on various design choices, we perform ablation studies on each component of \ourmethod{}. All experiments in this section are conducted on the AutoFormer search space and the CIFAR-100 dataset. 
It can be observed from Table~\ref{tab:beta} that both $\mathcal{M}_{s}$ and $\mathcal{M}_{t}$  are important to achieve remarkable rank consistency. 
We also explore different combinations of weight parameters $\alpha$ and $ \beta $ in Equation~\ref{eq:tvt}. In \ourmethod{}, $ \alpha $ and $ \beta $ play the role of balancing teacher-based and self-based information. The experimental results in Table~\ref{tab:beta} demonstrate that \ourmethod{} can achieve excellent performance with a suitable setting of weight parameters.
Moreover, we replace the spatial attention function in $\mathcal{M}_{t}$ by two different transformation methods, including the Sample Relation~\cite{sp} and MMD Metric~\cite{nst}. To validate the effectiveness of different transformation functions, we implement each method on \ourmethod{}-$\mathcal{M}_{t}$. The experimental results in Table~\ref{tab:transform} show that the attention map function brings the best performance.

\section{Conclusion}
This paper introduces a training-free ViT search framework on tiny datasets. 
Unlike existing training-free methods, our TVT is the first work focusing on searching ViTs on tiny datasets for distilling with the given ConvNet Teacher. Specifically, we discover the failures of existing training-free methods and present a novel zero-cost proxy with a teacher-aware metric and student-capability metric. 
Based on the proposed zero-cost proxy, we conduct a student ViT architecture search in a training-free manner, achieving significant accuracy gains. Extensive experiments validate the efficiency and effectiveness of TVT in various tiny datasets and search spaces. We hope this elegant and practical approach will inspire more investigation into the broader application of transformers on tiny datasets.

\bibliographystyle{IEEEbib}
\bibliography{strings,refs}

\begin{thebibliography}{10}

\bibitem{localvit}
Yawei Li, K.~Zhang, Jie Cao, Radu Timofte, and Luc~Van Gool,
\newblock ``Localvit: Bringing locality to vision transformers,''
\newblock {\em ArXiv}, vol. abs/2104.05707, 2021.

\bibitem{NWOT}
Joe Mellor, Jack Turner, Amos Storkey, and Elliot~J Crowley,
\newblock ``Neural architecture search without training,''
\newblock in {\em ICML}, 2021.

\bibitem{DSS}
Qinqin Zhou, Kekai Sheng, Xiawu Zheng, Ke~Li, Xing Sun, Yonghong Tian, Jie
  Chen, and Rongrong Ji,
\newblock ``Training-free transformer architecture search,''
\newblock in {\em Proceedings of the IEEE/CVF Conference on Computer Vision and
  Pattern Recognition}, 2022, pp. 10894--10903.

\bibitem{deit}
Hugo Touvron, Matthieu Cord, Matthijs Douze, Francisco Massa, Alexandre
  Sablayrolles, and Herv{\'e} J{\'e}gou,
\newblock ``Training data-efficient image transformers \& distillation through
  attention,''
\newblock in {\em International Conference on Machine Learning}, 2021.

\bibitem{Jia2021EfficientVT}
Ding Jia, Kai Han, Yunhe Wang, Yehui Tang, Jianyuan Guo, Chao Zhang, and
  Dacheng Tao,
\newblock ``Efficient vision transformers via fine-grained manifold
  distillation,''
\newblock {\em ArXiv}, vol. abs/2107.01378, 2021.

\bibitem{Wu2022TinyViTFP}
Kan Wu, Jinnian Zhang, Houwen Peng, Mengchen Liu, Bin Xiao, Jianlong Fu, and
  Lu~Yuan,
\newblock ``Tinyvit: Fast pretraining distillation for small vision
  transformers,''
\newblock {\em ArXiv}, vol. abs/2207.10666, 2022.

\bibitem{liu2023norm}
Xiaolong Liu, Lujun Li, Chao Li, and Anbang Yao,
\newblock ``Norm: Knowledge distillation via n-to-one representation
  matching,''
\newblock {\em arXiv preprint arXiv:2305.13803}, 2023.

\bibitem{li2023auto}
Lujun Li, Peijie Dong, Zimian Wei, and Ya~Yang,
\newblock ``Automated knowledge distillation via monte carlo tree search,''
\newblock in {\em ICCV}, 2023.

\bibitem{lishadow}
Lujun Li and Zhe Jin,
\newblock ``Shadow knowledge distillation: Bridging offline and online
  knowledge transfer,''
\newblock in {\em NeuIPS}, 2022.

\bibitem{li2023kd}
Lujun Li, Peijie Dong, Anggeng Li, Zimian Wei, and Yang Ya,
\newblock ``Kd-zero: Evolving knowledge distiller for any teacher-student
  pairs,''
\newblock in {\em Thirty-seventh Conference on Neural Information Processing
  Systems}, 2023.

\bibitem{li2022tf}
Lujun Li, Liang Shiuan-Ni, Ya~Yang, and Zhe Jin,
\newblock ``Teacher-free distillation via regularizing intermediate
  representation,''
\newblock in {\em IJCNN}, 2022.

\bibitem{li2022SFF}
Lujun Li, Liang Shiuan-Ni, Ya~Yang, and Zhe Jin,
\newblock ``Boosting online feature transfer via separable feature fusion.,''
\newblock in {\em IJCNN}, 2022.

\bibitem{li2020explicit}
Lujun Li, Yikai Wang, Anbang Yao, Yi~Qian, Xiao Zhou, and Ke~He,
\newblock ``Explicit connection distillation,''
\newblock 2020.

\bibitem{li2022self}
Lujun Li,
\newblock ``Self-regulated feature learning via teacher-free feature
  distillation,''
\newblock in {\em ECCV}, 2022.

\bibitem{shao2023catch}
Shitong Shao, Xu~Dai, Shouyi Yin, Lujun Li, Huanran Chen, and Yang Hu,
\newblock ``Catch-up distillation: You only need to train once for accelerating
  sampling,''
\newblock {\em arXiv preprint arXiv:2305.10769}, 2023.

\bibitem{li2022locality}
Kehan Li, Runyi Yu, Zhennan Wang, Li~Yuan, Guoli Song, and Jie Chen,
\newblock ``Locality guidance for improving vision transformers on tiny
  datasets,''
\newblock in {\em European Conference on Computer Vision}. Springer, 2022, pp.
  110--127.

\bibitem{wei2023Auto-Prox}
Zimian Wei, Lujun Li, Peijie Dong, , Anggeng Li, Menglong Lu, Hengyue Pan, and
  Dongsheng Li,
\newblock ``Auto-prox: Training-free vision transformer architecture search via
  automatic proxy discovery,''
\newblock {\em AAAI}, 2024.

\bibitem{li2021nas}
Yiming Hu, Xingang Wang, Lujun Li, and Qingyi Gu,
\newblock ``Improving one-shot nas with shrinking-and-expanding supernet,''
\newblock {\em Pattern Recognition}, 2021.

\bibitem{linas2}
Peijie Dong, Xin Niu, Lujun Li, Linzhen Xie, Wenbin Zou, Tian Ye, Zimian Wei,
  and Hengyue Pan,
\newblock ``Prior-guided one-shot neural architecture search,''
\newblock {\em arXiv preprint arXiv:2206.13329}, 2022.

\bibitem{lichengp}
Kunlong Chen, Liu Yang, Yitian Chen, Kunjin Chen, Yidan Xu, and Lujun Li,
\newblock ``Gp-nas-ensemble: a model for the nas performance prediction,''
\newblock in {\em CVPRW}, 2022.

\bibitem{dong2023diswot}
Peijie Dong, Lujun Li, and Zimian Wei,
\newblock ``Diswot: Student architecture search for distillation without
  training,''
\newblock in {\em CVPR}, 2023.

\bibitem{dong2023emq}
Peijie Dong, Lujun Li, Zimian Wei, Xin Niu, Zhiliang Tian, and Hengyue Pan,
\newblock ``Emq: Evolving training-free proxies for automated mixed precision
  quantization,''
\newblock {\em arXiv preprint arXiv:2307.10554}, 2023.

\bibitem{syflow}
Hidenori Tanaka, Daniel Kunin, Daniel~L Yamins, and Surya Ganguli,
\newblock ``Pruning neural networks without any data by iteratively conserving
  synaptic flow,''
\newblock {\em NeurIPS}, 2020.

\bibitem{TENAS}
Wuyang Chen, Xinyu Gong, and Zhangyang Wang,
\newblock ``Neural architecture search on imagenet in four gpu hours: A
  theoretically inspired perspective,''
\newblock in {\em ICLR}, 2020.

\bibitem{EcoNAS}
D.~Zhou, Xinchi Zhou, Wenwei Zhang, Chen~Change Loy, Shuai Yi, Xuesen Zhang,
  and Wanli Ouyang,
\newblock ``Econas: Finding proxies for economical neural architecture
  search,''
\newblock {\em 2020 IEEE/CVF Conference on Computer Vision and Pattern
  Recognition (CVPR)}, 2020.

\bibitem{ENAS}
Hieu Pham, Melody~Y. Guan, Barret Zoph, Quoc~V. Le, and Jeff Dean,
\newblock ``Efficient neural architecture search via parameter sharing,''
\newblock in {\em ICML}, 2018.

\bibitem{zhu2021hard}
Chuang Zhu, Wenkai Chen, Ting Peng, Ying Wang, and Mulan Jin,
\newblock ``Hard sample aware noise robust learning for histopathology image
  classification,''
\newblock {\em IEEE Transactions on Medical Imaging}, vol. 41, no. 4, pp.
  881--894, 2021.

\bibitem{zeronas}
Mohamed~S Abdelfattah, Abhinav Mehrotra, {\L}ukasz Dudziak, and Nicholas~Donald
  Lane,
\newblock ``Zero-cost proxies for lightweight nas,''
\newblock in {\em ICLR}, 2020.

\bibitem{snip}
Namhoon Lee, Thalaiyasingam Ajanthan, and Philip~HS Torr,
\newblock ``Snip: Single-shot network pruning based on connection
  sensitivity,''
\newblock {\em arXiv preprint arXiv:1810.02340}, 2018.

\bibitem{chen2021autoformer}
Minghao Chen, Houwen Peng, Jianlong Fu, and Haibin Ling,
\newblock ``Autoformer: Searching transformers for visual recognition,''
\newblock in {\em Proceedings of the IEEE/CVF International Conference on
  Computer Vision}, 2021, pp. 12270--12280.

\bibitem{grasp}
Chaoqi Wang, Guodong Zhang, and Roger Grosse,
\newblock ``Picking winning tickets before training by preserving gradient
  flow,''
\newblock {\em arXiv preprint arXiv:2002.07376}, 2020.

\bibitem{sp}
Frederick Tung and Greg Mori,
\newblock ``Similarity-preserving knowledge distillation,''
\newblock in {\em ICCV}, 2019.

\bibitem{nst}
Zehao Huang and Naiyan Wang,
\newblock ``Like what you like: {K}nowledge distill via neuron selectivity
  transfer,''
\newblock {\em arXiv:1707.01219}, 2017.

\end{thebibliography}

\end{document}